\documentclass[10pt,letterpaper]{article}

\usepackage{cogsci}

\cogscifinalcopy 

\usepackage{graphicx}
\usepackage{pslatex}
\usepackage{apacite}
\usepackage{float} 

\title{A Model of Fast Concept Inference with Object-Factorized Cognitive Programs}
 
\author{
{\large \bf Daniel P. Sawyer (dsawyer@caltech.edu)} \\
  Biology and Bioengineering, California Institute of Technology, Pasadena, CA, USA 
  \AND {\large \bf Miguel L\'azaro-Gredilla (miguel@vicarious.com)} \\
  Vicarious AI, San Francisco, USA
  \AND {\large \bf Dileep George (dileep@vicarious.com)} \\
  Vicarious AI, San Francisco, USA
  }

\begin{document}

\maketitle

\begin{abstract}
  The ability of humans to quickly identify general concepts from a handful of images has proven difficult to emulate with robots. Recently, a computer architecture was developed that allows robots to mimic some aspects of this human ability by modeling concepts as cognitive programs using an instruction set of primitive cognitive functions. This allowed a robot to emulate human imagination by simulating candidate programs in a world model before generalizing to the physical world. However, this model used a naive search algorithm that required 30 minutes to discover a single concept, and became intractable for programs with more than 20 instructions. To circumvents this bottleneck, we present an algorithm that emulates the human cognitive heuristics of object factorization and sub-goaling, allowing human-level inference speed, improving accuracy, and making the output more explainable.

\end{abstract}

\section{Introduction}
\begin{figure}[t]
\begin{center}
  \includegraphics[width=0.48\textwidth]{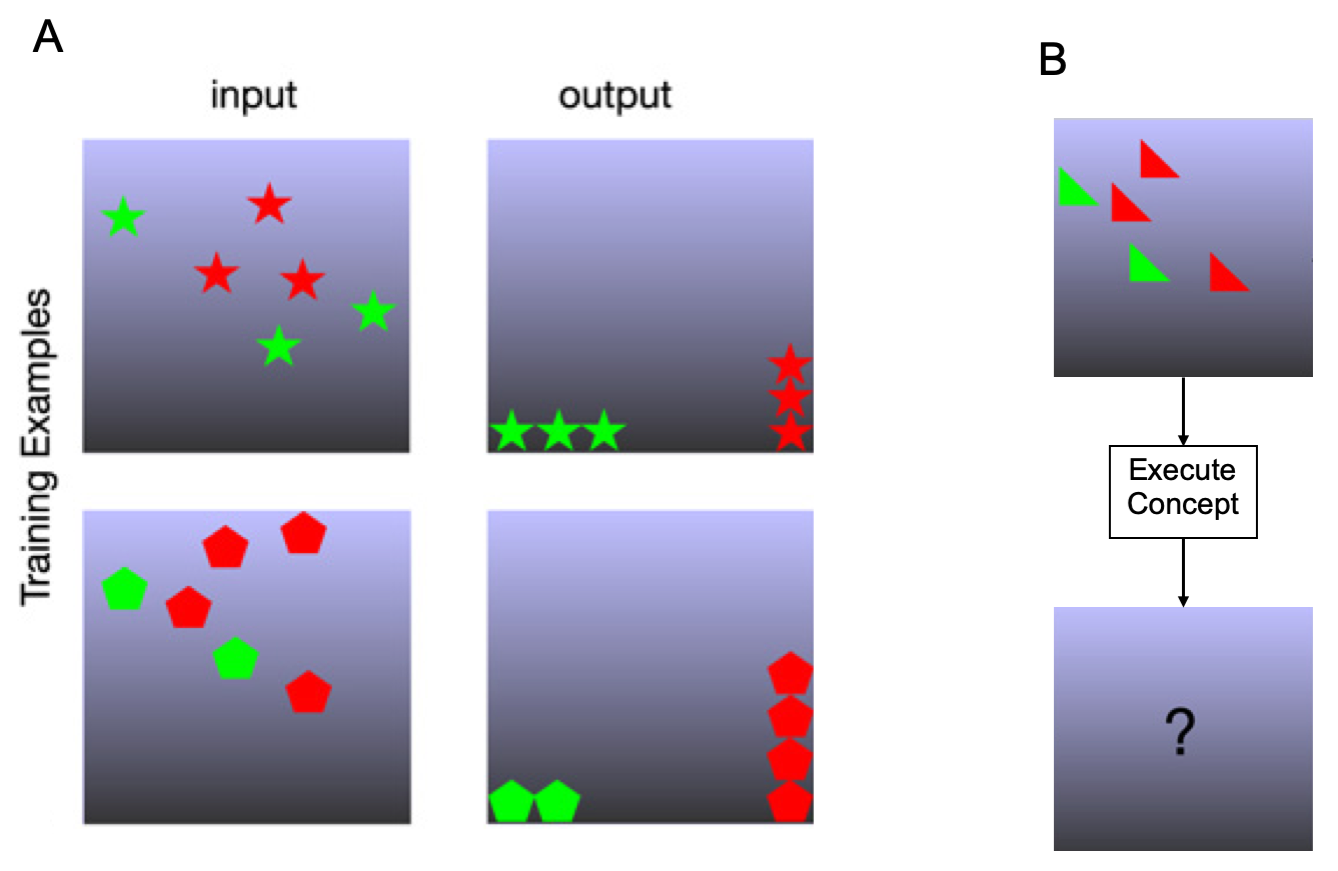}
\end{center}
\caption{People can easily understand the concept conveyed in pairs of images, a capability that is exploited by LEGO and IKEA assembly diagrams. (\textbf{A}) People interpret the concept conveyed by these images as stacking red objects vertically on the right and green objects horizontally at the bottom. (\textbf{B}) Given a novel image, people can predict what the result of executing the concept would be. Adapted with permission from \citeA{lazaro-gredilla_beyond_2019}.}
\label{fig:summary}
\end{figure}

Humans can readily infer the high-level concept represented in a pair of images and then apply it in a diverse array of circumstances (Fig.\ref{fig:summary}A-B). This capability allows everything from LEGO instructions to traffic signs to provide language-independent guides to human behavior. Robots, in contrast, are typically programmed by tediously specifying a sequence of movements or poses for a single, highly controlled setting. More recently, imitation learning has been employed in attempts to provide more versatility to robots by allowing them to learn from demonstrations \cite{akgun_keyframe-based_2012, duan_one-shot_2017}. However, by using fragile mapping from image frame pixels to actions, imitation learning policies often fail to generalize in response to changes in object appearance or lighting conditions \cite{tung_reward_2018}.

Providing robots the ability to infer concepts with the speed and data efficiency of humans would not only allow broader task automation, it would make human-robot communication more intuitive, successful task completion more explainable, and failures more readily diagnosed.

\begin{figure}[t]
\begin{center}
  \includegraphics[width=0.48\textwidth]{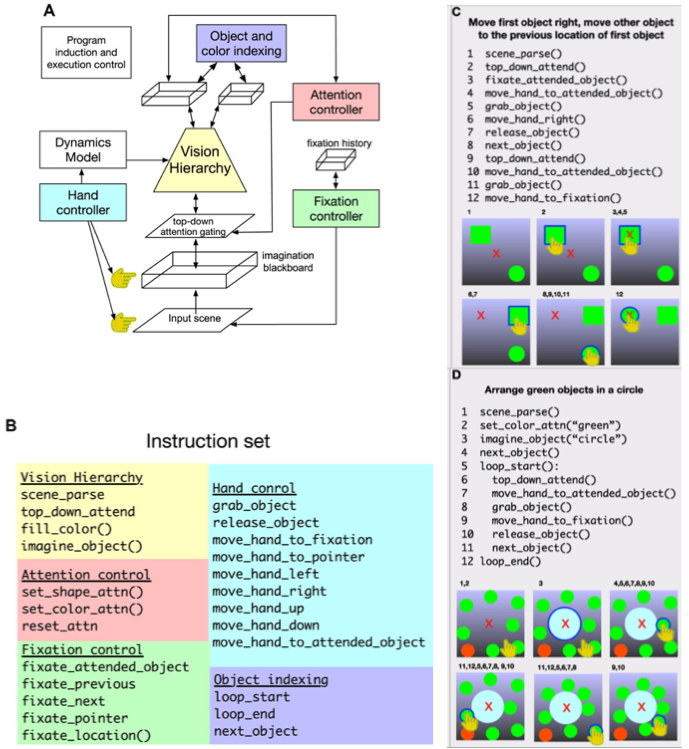}
\end{center}
\caption{VCC architecture and program execution examples. (\textbf{A}) Functional components of VCC and their interactions. The vision hierarchy can parse the input scene and identify, attend to, and imagine objects. The hand controller moves the hand to different locations in the scene, and the fixation controller commands position the center of the eye. Object indexing commands iterate through the objects currently attended to. The attention controller can set the current attention based on object shape or color. (\textbf{B}) The full instruction set of VCC. Parentheses denote instructions with arguments. (\textbf{C-D}) examples of discovered programs and visualizations of their execution steps. Digits next to the visualizations correspond to program line numbers. Blue highlight indicates the shape attended to. Red ``X'' indicates the point of fixation. Adapted with permission from \citeA{lazaro-gredilla_beyond_2019}.}
\label{fig:architecture}
\end{figure}

Recently, an architecture called visual cognitive computer (VCC) was developed that allowed robots to learn concepts from fewer than 10 input/output image pairs and then apply them in diverse physical settings \cite{lazaro-gredilla_beyond_2019}. VCC is based on key ideas from cognitive science, including image schemas \cite{mandler_defining_2014}, deictic mechanisms \cite{ballard_deictic_1997}, perceptual symbol systems \cite{barsalou_perceptual_1999}, visual routines \cite{ullman_object_1996}, and mental imagery \cite{roelfsema_early_2016}. The central idea was to represent concepts as cognitive programs: sequences of primitive instructions analogous to the operations in the instruction set of a microprocessor. Rather than basic arithmetic and control-flow functions, the VCC instructions implement fundamental cognitive functions such as parsing a visual scene, directing gaze and attention, imagining new objects, manipulating the contents of a visual working memory, and controlling arm movement (Fig.\ref{fig:architecture}A-B).

Because concepts are abstract in nature, the VCC must parse pixel-based input scenes into symbolic lists of objects and their visuospatial properties, such as position, shape, and color. However, to simulate non-symbolic interactions such as object collisions, the VCC must also be capable of mapping a symbolic scene representations back to the pixel level. VCC accomplishes both of these functions through a vision hierarchy (VH) based on a generative model that achieves near human-level performance on image segmentation tasks \cite{george_generative_2017} and reproduces visual cortex phenomena \cite{lavin_explaining_2018} with computational requirements compatible with the anatomical constraints of cortical microcircuits \cite{george_cortical_2018}. This architecture allows a robot to learn concepts before any interaction with the physical world by simulating the results of candidate programs on an imagination blackboard, which serves a function similar to the visual cortex \cite{roelfsema_early_2016}.

Prior to program induction, a given image is processed as follows. The VH takes as input an RGB image and outputs a list of objects, where each object is represented as a $1\times 5$ vector of features encoding size, horizontal position, vertical position, shape, and color. Size and position take continuous values while shape and color take categorical values (coded as integers). The feature spaces for shape and color are $\{$`square', `circle', `line', `diamond', `triangle', `star'$\}$ and $\{$`red', `green', `blue', `yellow'$\}$, respectively.

Each instruction can be thought of as an operation that changes the state of the agent (hand position, fixation position, contents of attention buffer, etc.) or its environment (color and position properties of objects). For a given concept, the general program induction problem is to find a sequence of instructions belonging to an instruction set of size $N$ that successfully changes the properties of the objects in every input example to match the objects in every output example (Fig.\ref{fig:architecture}C-D).

More formally, each processed scene $S$ containing $k$ objects may be represented as a set of object vectors $S = \{o_1, \ldots, o_k\}$, and each program as a function $p$ that transforms an input scene $S_I$ into a new scene $p(S_I)$ with the same number of elements each in the same feature space. An object $o_i$ in a transformed input scene $p(S_I)$ is said to be ``matched'' with respect to a target output scene $S_T$ if there exists an object $o_j$ in $S_T$ such that $o_i = o_j$. A program $p$ is said to have solved an input/output example $(S_I, S_T)$ if every object in $p(S_I)$ is matched. There are some caveats to this formal description, an exhaustive discussion of which is beyond of the scope of this paper. Please refer to the supplementary material of \citeA{lazaro-gredilla_beyond_2019} for details.

While the VCC represents an impressive synthesis of cognitive science principles in an architecture with the novel capability of discovering conceptual understanding without demonstration data, its ability to scale to more complex tasks is limited by the nature of its program induction algorithm. Apart from the input and output example images, the only information available to the VCC is whether a given program solves every example of a concept, which makes program induction a tree search problem, where each program is a node on an $N$-ary tree. If we assume that a given concept can be solved by a unique ground-truth program of length $L$, the brute-force run time is $O(N^L)$, which quickly becomes intractable for longer programs. The search algorithm described in \citeA{lazaro-gredilla_beyond_2019} employs several heuristics to make the task more tractable by allowing the VCC to learn from previous attempts:
\begin{itemize}
    \item Pruning of invalid nodes (programs) from further search
    \item Argument prediction using fixation guidance and convolutional neural networks (CNNs)
    \item A Markov model that prioritizes search by learning instruction transition probabilities that are stored in a Markov transition matrix\footnote{Entry $(i,j)$ of the Markov transition matrix represents the probability that instruction $i$ follows instruction $j$ in a program.}.
    \item Identification of subroutines
\end{itemize}

Pruning of invalid programs was the most effective heuristic, since, typically, over 50\% of instruction transitions were never valid (e.g. calling \verb`release_object` before \verb`grab_object`). 

Argument prediction effectively reduced $N$ by 33\% in the best case since the instruction set contains 36 primitives if counting different argument options as separate instructions, but only 24 otherwise. This benefit is mostly orthogonal to that of pruning since the validity of a transition is mostly independent of an instruction's arguments. 

The Markov model also acted by effectively reducing $N$, though its benefits overlap with those of pruning and argument prediction. 

These heuristics dramatically improve the speed of the search, but they do not fundamentally change the exponential scaling of the run time because they do not change the minimum depth $L$ of the search tree. The use of subroutines identified from common sequences of instructions is the only heuristic that, in theory, effectively reduces $L$. However, this approach did not empirically result in significant improvement.

These heuristics allowed the VCC to solve most of the $546$ concepts on which it was tested with reasonable efficiency, solving a typical concept in $30$ minutes after executing $3$ million programs. However, it was unable to solve many of the more complex concepts, including most concepts similar to that illustrated in Fig.\ref{fig:summary} that involve stacking variable numbers of two different types of objects. This is especially concerning for future applications that will require larger instruction sets and longer programs. Moreover, because success of a program is a binary metric, failure cases include no partial solutions to aid in explaining the reason for the failure.

Here, we address this bottleneck with a novel program induction algorithm using a divide-and-conquer approach. We provide the VCC with more fine-grained feedback during the search by evaluating program success on the basis of individual objects rather than the entire scene. In combination with a program mutation function that addresses multi-object dependencies, this approach greatly improves the scalability of the VCC and brings it closer to human performance in terms of inference time and explainability.

\section{Fast Program Induction Algorithm}
Although the VCC \cite{lazaro-gredilla_beyond_2019} used object-factorized representations for parsing, dynamics, and instruction set, its search algorithm had two primary deficiencies. First, it did not exploit the object-factorizations in the search process. Second, the search was a purely feed-forward open-loop process, where partial attainments of goals did not alter the search process. That is, the search was driven purely by the input image, and the output-image was used purely for verification. In contrast, humans use object-factorizations and sub-goaling to drive the search process, and the sub-goals are obtained by jointly considering the input and output images.

\subsection{Object Factorization and Sub-goaling}
The central idea of our new approach is to factorize the search by object. That is, split the input/output examples into sub-goals in which we search for a program that transforms the properties (usually spatial coordinates and sometimes color) of a single object in the input to match the output. Assuming each unmatched object requires the same number of instructions to solve, this approach will solve a concept with $k$ unmatched objects in $O(kN^{L/k})$ time as long as the object sub-goals are independent.

There are several ways to incorporate this approach as a heuristic in the program induction process. The most important choices are how to formulate and schedule the sub-goals for search. As an example, suppose we have 3 unmatched objects and we represent the solved state of each with a one-hot array $[m_1\; m_2\; m_3],\: m_i \in \{0,1\}$. One strategy would be to choose a random ordering of objects to match and try to solve them sequentially in that order. For example, one ordering could be $[1\; 0\; 0],\: [1\; 1\; 0],\: [1\; 1\; 1]$. After a program is found for a sub-goal, we restart the search with that program as the root node. If, after trying for some threshold number of programs, we are not able to solve the concept with a particular ordering, we restart from scratch with another ordering. However, this strategy does not addresses how to identify conceptually equivalent objects from separate examples. Consider all the concepts that require moving one object to the left. The defining property of the object to be moved left could be color, shape, distance from the center, etc. This is a nontrivial unsupervised clustering problem, and a single error could render a concept unsolvable because the exact clustering is used as the basis for evaluating the success of a program on a given sub-goal.

For this reason, we opted for the following strategy: run the search as normal until a program matches at least one previously-unmatched object in every example, then restart the search with that program as the root node. The fact that each object from a different example is matched by the same program is the best evidence that they belong to the same conceptual cluster. Conversely, the fact that a program matches an object in every example is evidence that it represents a valid sub-concept.

\subsection{Program Mutation}
There are some concepts that fundamentally cannot be solved using an object factorization approach on its own. For instance, the concepts that have different numbers of unmatched objects in each example require one or more loops to solve. However, the object factorization strategy is unlikely to find programs with loops. This is because the \verb#loop_start# instruction must occur prior to the sequence of instructions that matches each individual object (Fig.\ref{fig:architecture}D), but the object factorization fixes a successful instruction sequence in the new root node, which precludes insertion of the necessary  \verb#loop_start#. Another case is when an object must be moved to the previous location of a different object (Fig.\ref{fig:architecture}C). Here, the VCC must fixate the first object before it moves it. Otherwise, it will have no memory of the location.

To address these limitations while maintaining the benefits of object factorization, we introduce a program mutation function inspired by the technique of iterative mutation, screening, and selection in protein engineering known as directed evolution \cite{arnold_design_1998}. The mutation function finds all unseen, valid, single-instruction transpositions, deletions, changes, and insertions (except those that would change the first or last instruction) for an input program of length $l$ and returns the corresponding ``mutant'' programs that find at least as many objects as the input program. Any programs deemed valid (defined as having a finite description length calculated from the log Markov transition matrix) are executed on the VCC and are counted toward the total program budget. Because a program must have loop closure to solve any objects, if an inserted instruction is \verb#loop_start#, then we also try to insert a \verb#loop_end# instruction at each position after that. 

The mutation function is used as follows. A mutation threshold is initialized as $n_{\mathrm{progs}}/10$ where $n_{\mathrm{progs}}$ is the program budget. If at least one sub-goal has been solved and the number of visited programs is greater than the threshold, the mutation function is executed on the current root node. If any mutants solve all the objects, the shortest mutant is the found program. If any mutants find new (but not all) objects in all examples, they become the new root nodes for the next sub-goal search. If any mutants find the same objects, they are pushed onto the search queue. Otherwise, the search continues with no change.

\subsection{Other Improvements}
To further narrow the search space and to prevent the VCC from inadvertently undoing its progress after achieving a sub-goal, we enforced that a program should be pruned from further search if it results in a previously-matched object becoming unmatched.

We also enforced that an object must be released from the hand in order to qualify as matched. Previously, an object could qualify as matched if in the correct position but still in the hand, which introduced an added dependency between sub-goals since the VCC would need to release the grabbed object before attending to the next one. This makes some programs slightly more difficult by requiring the additional \verb#release_object# instruction at the end, but overall it improved performance in our framework.

\section{Results}
\subsection{Performance Improvements}
We used Dijkstra's algorithm for search, where the ``distance'' to a given program is its description length calculated from the Markov transition matrix. We did not use the more-popular A* search algorithm because it requires a consistent estimate of the distance to the  goal, which is non-trivial in our problem setting\footnote{Unlike in physical shortest-path problems, it is not clear how to estimate the remaining distance (description length). For example, multiplying the number of unmatched objects by the average description length necessary to match an object would not be a consistent estimate since an arbitrary number of objects in some concepts can be matched using the loop instructions.}. The search used 10.4 GB of memory. The CNN and provided fixation were used for argument prediction, and the first-order Markov model was trained using the 16 ground truth programs of length 6 or less. Subroutines were disabled. The search was executed on a laptop using 8 cores.

To speed up testing, the transition matrix was initialized with an empirical dependency graph that sets to zero the probability of any transition that is never needed to solve any of the concepts. This approximates the speed improvement obtained from the Markov model after running the search for multiple iterations.

\begin{table}[t]
\begin{center} 
\caption{Performance improvements.}
\vskip 0.12in
\begin{tabular}{lll} 
\hline
                   & Naive search & Object factorization \\
\hline
    Concepts found & $526\: (96.3\%)$   & $534\: (97.8\%)$   \\
    Program budget & $3,000,000$        & $4,000$ \\
    Median run time& $30$ min/concept   & $1$ min/concept  \\
\hline
\end{tabular} 
\end{center} 
\label{table:performance}
\end{table}

Compared to the naive search, our object factorization algorithm reduced the failure rate from 3.7\% to 2.2\% using a program search budget three orders of magnitude smaller (Table \ref{table:performance}).

\begin{figure}[t]
\begin{center}
\includegraphics[width=0.5\textwidth]{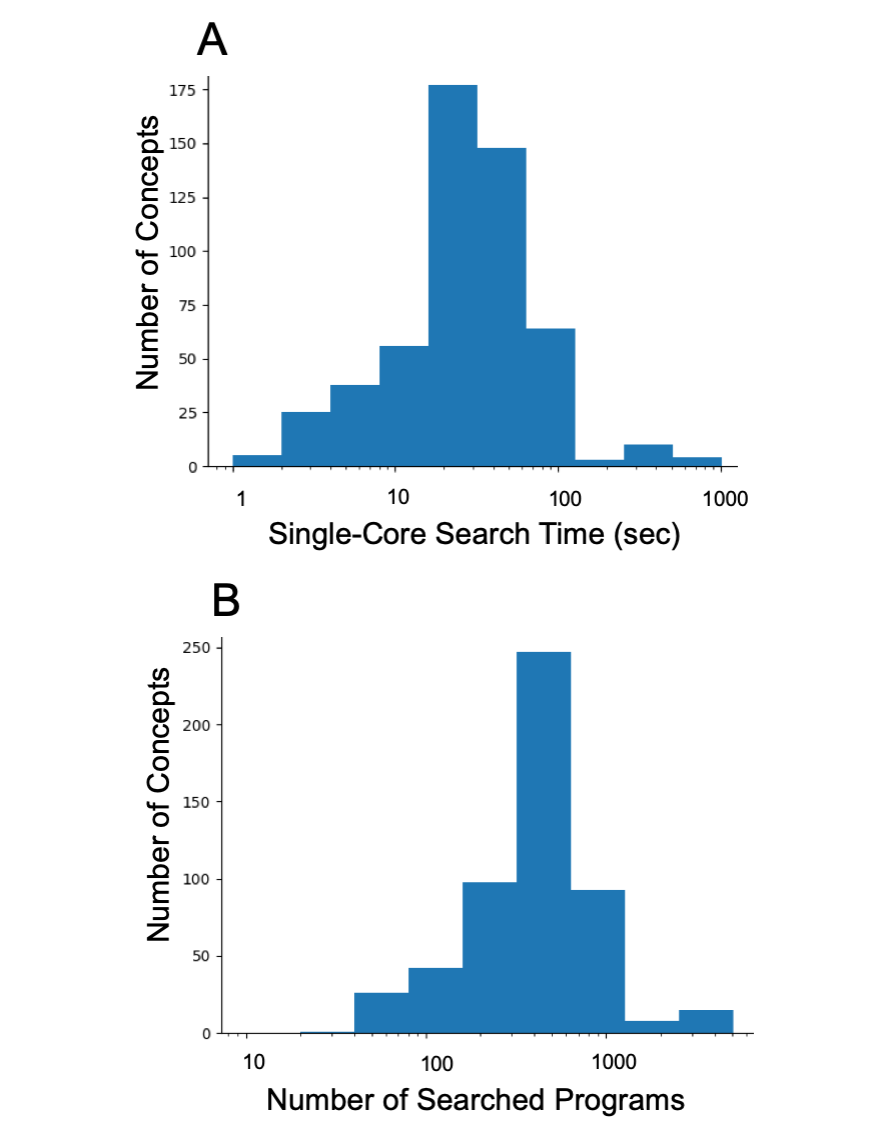}
\end{center}
\caption{Distribution of run times (\textbf{A}) and visited programs (\textbf{B}) for object factorization search with sub-goals.}
\label{fig:performance}
\end{figure}

The histograms in Fig.\ref{fig:performance} show the distribution of the number of visited programs and per-concept search time for the object factorization search algorithm. 95\% of concepts are found in under 2 minutes, with a median per-concept search time of 1 min. This is similar to the time typically needed by humans to identify simple visual concepts, such those in the Raven Progressive Matrices Test, where subjects have 40 minutes to identify 46 concepts \cite{carpenter_what_1990}.\\

Though this relative reduction in run time is significant, it is more than an order of magnitude smaller than the relative reduction in program budget. As we made no attempt to optimize our code for speed, we suspect this discrepancy is due to the more thorough optimization of search algorithm code in \citeA{lazaro-gredilla_beyond_2019}.

Moreover, these gains likely understate the improvement offered by object factorization because they do not consider performance improvements on concepts involving more than two functionally independent types of objects. For instance, based on the median run time, a concept requiring 8 objects to be moved to the 4 edges and 4 corners could be discovered by object factorization search in under 10 minutes but would be exponentially more difficult for the naive search, requiring at least 2 days.

\subsection{Explainability Improvements}
As the internal mechanisms of both the VCC architecture and our object factorization search algorithm were inspired by models of human cognition \cite{mandler_defining_2014, ballard_deictic_1997, barsalou_perceptual_1999, ullman_object_1996, roelfsema_early_2016}, we should expect its performance to be more intuitive and explainable for humans than alternative approaches such as neural networks. We demonstrate this enhanced explainability by analyzing failure cases and showing how the reasons for failure can be readily determined from the algorithm's output. Here, the sub-goaling of object factorization allows us to examine any partial solutions generated by the VCC to diagnose the reason for failure. Similarly, the use of instructions based on cognitive primitives in the VCC architecture makes such output readily interpretable. We leave to future work the question as to whether the implementation of concept inference as cognitive programs is an accurate model of human cognition.

The new search algorithm failed to find 12 of the 546 concepts. Among the 12 failure cases, there appear to be 4 reasons the VCC fails to find the correct program: solving the objects in the wrong order (2 cases), mistaken object identity (6 cases), faulty argument prediction (2 cases), and lack of sufficient search budget (2 cases). We provide an example of each failure mode below (we abbreviate \verb#move_hand_to_attended_object# as \verb#move_hand_to_object# here for space).

\begin{figure}[t]
\begin{center}
\includegraphics[width=0.5\textwidth]{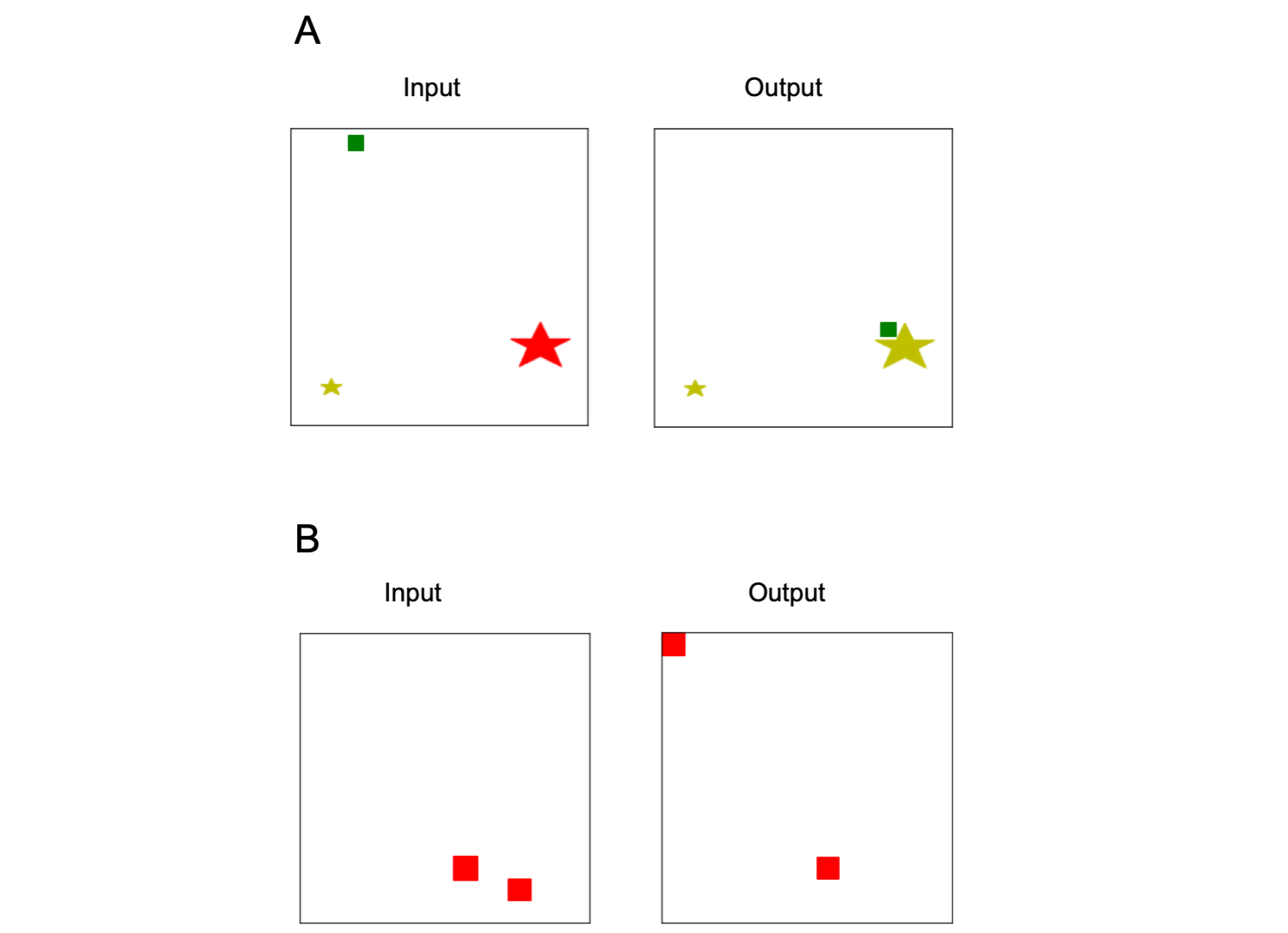}
\end{center}
\caption{Input/output example images for failure cases of wrong object order (\textbf{A}) and mistaken object identity (\textbf{B}).}
\label{fig:failures}
\end{figure}

\begin{table}[H]
\begin{center} 
\caption{Failure case: wrong object order\\ 
Concept: make the green object touch the red object and change the color of the object being touched} 
\label{table:wrong_order}
\vskip 0.12in
\begin{tabular}{ll}
\hline
Best found program & Ground truth program \\
\hline
\verb#scene_parse#&\verb#scene_parse# \\
\verb#set_color_attn(red)#&\verb#set_color_attn(green)# \\
\verb#top_down_attend#&\verb#top_down_attend# \\
\verb#fill_color(yellow)#&\verb#move_hand_to_object# \\
                        & \verb#grab_object# \\
                        & \verb#reset_attn# \\
                        & \verb#top_down_attend# \\
                        & \verb#move_hand_to_object# \\
                        & \verb#release_object# \\
                        & \verb#fill_color(yellow)# \\
\hline
\end{tabular} 
\end{center} 
\end{table}

In the example of wrong object order in Table \ref{table:wrong_order}, changing the color of the red object requires a shorter program than moving the green object. The algorithm thus does this first, ``solving'' the object and resetting the root node (Fig.\ref{fig:failures}A). However, the color of the object was used to identify it as the object to be touched, meaning any argument prediction of the color to attend to will no longer be accurate. In this case, argument prediction likely assigned a very low probability to all \verb#set_color_attn()# instructions with arguments other than `green'. Failures in this category could likely be solved with a strategy that reserves some of the search budget for attempts at matching the objects in a different order.

\begin{table}[H]
\begin{center} 
\caption{Failure case: mistaken object identity\\ 
Concept: move the central objects to the upper left corner and move the other object to the previous object's location} 
\label{table:mistaken_id}
\vskip 0.12in
\begin{tabular}{ll}
\hline
Best found program & Ground truth program \\
\hline
\verb#scene_parse#&\verb#scene_parse# \\
\verb#top_down_attend#&\verb#top_down_attend# \\
\verb#fixate_object#&\verb#fixate_object# \\
\verb#move_hand_to_object#&\verb#move_hand_to_object# \\
\verb#grab_object#& \verb#grab_object# \\
\verb#move_hand_up#& \verb#move_hand_up# \\
                        & \verb#move_hand_left# \\
                        & \verb#release_object# \\
                        & \verb#next_object# \\
                        & \verb#top_down_attend# \\
                        & \verb#move_hand_to_object# \\
                        & \verb#grab_object# \\
                        & \verb#move_hand_to_fixation# \\
\hline
\end{tabular} 
\end{center} 
\end{table}

In the example of mistaken object identity in Table \ref{table:mistaken_id}, the two objects are both red squares with identical size, and so it appears that the object close to the center does not move since the other object is moved to its previous location (Fig.\ref{fig:failures}B). Consequently, when that object is moved up, the VCC considers it to be a matched object becoming unmatched and prunes the program from further search. Failures in this category, which account for half of the failure cases, could be solved by encoding an ``object ID'' in the input and output examples, or by relaxing the node pruning condition.

\begin{table}[H]
\begin{center} 
\caption{Failure case: faulty argument prediction\\ 
Concept: make the star shape touch the circle shape} 
\label{table:faulty_arg}
\vskip 0.12in
\begin{tabular}{ll}
\hline
Best found program & Ground truth program \\
\hline
\verb#scene_parse#&\verb#scene_parse# \\
\verb##&\verb#set_shape_attn(star)# \\
\verb##&\verb#top_down_attend# \\
\verb##&\verb#move_hand_to_object# \\
       & \verb#grab_object# \\
       & \verb#reset_attn# \\
       & \verb#set_shape_attn(circle)# \\
       & \verb#top_down_attend# \\
       & \verb#move_hand_to_object# \\
       & \verb#release_object# \\
\hline
\end{tabular} 
\end{center} 
\end{table}

In the example of faulty argument prediction in Table \ref{table:faulty_arg}, an extremely low probability was assigned to the (correct) `star' argument, causing all transitions to \verb#set_shape_attn(star)# to be assigned an extremely low value in the instruction transition matrix, so the correct node was never be visited. Because argument prediction was implemented with neural networks, we are unable to explain why it failed in this particular case. This case serves as an example of the limitations in explainability imposed by the use of black-box models. Failures in this category could likely be solved with an improved argument prediction model. 

\begin{table}[H]
\begin{center} 
\caption{Failure case: insufficient search budget\\ 
Concept: swap locations} 
\label{table:low_budget}
\vskip 0.12in
\begin{tabular}{ll}
\hline
Best found program & Ground truth program \\
\hline
\verb#scene_parse#&\verb#scene_parse# \\
\verb##&\verb#top_down_attend# \\
\verb##&\verb#move_hand_to_object# \\
       & \verb#grab_object# \\
       & \verb#fixate_object# \\
       & \verb#move_hand_down# \\
       & \verb#release_object# \\
       & \verb#next_object# \\
       & \verb#top_down_attend# \\
       & \verb#move_hand_to_object# \\
       & \verb#grab_object# \\
       & \verb#fixate_object# \\
       & \verb#fixate_previous# \\
       & \verb#move_hand_to_fixation# \\
       & \verb#release_object# \\
       & \verb#reset_attention# \\
       & \verb#next_object# \\
       & \verb#top_down_attend# \\
       & \verb#move_hand_to_object# \\
       & \verb#grab_object# \\
       & \verb#fixate_next# \\
       & \verb#move_hand_to_fixation# \\
       & \verb#release_object# \\
\hline
\end{tabular} 
\end{center} 
\end{table}

Table \ref{table:low_budget} shows the most difficult concept in the data set. In addition to requiring $23$ instructions (the longest program requires $24$), it makes use of instructions such as \verb#fixate_previous# and \verb#fixate_next# that are rare in other concepts and thus are assigned low transition probabilities. Most importantly from the perspective of object factorization, this concept requires $15$ instructions before the first object is matched. Since the naive search algorithm was able to find all programs of length 16 with a search budget of $500,000$ programs, we might expect object factorization to overcome this failure and others like it if given a search budget in this range.

\section{Conclusions}
Overall, the object factorization approach and other improvements made to the search algorithm increased the search efficiency of the program induction by three orders of magnitude while also significantly decreasing the failure rate. Analysis of the few failure cases, aided by the improved explainability afforded by sub-goaling, suggests the new failure rate can be at least halved with minor changes to the object identification or pruning strategy. Run time can likely also be further improved, as the current implementation is not optimized for computational efficiency. This dramatic reduction in the computational cost of concept inference opens the door to several future directions of exploration, such as extending the instruction set to solve concepts in 3D space or solving more complex compositional concepts requiring hierarchies of sub-goals. Such developments will bring us closer to robots that learn tasks from diagrams on the fly with human-like flexibility.

\bibliographystyle{apacite}

\setlength{\bibleftmargin}{.125in}
\setlength{\bibindent}{-\bibleftmargin}

\bibliography{ObjFactorization}

\end{document}